\documentclass[sigconf,nonacm]{acmart}
\renewcommand\footnotetextcopyrightpermission[1]{} % removes footnote with conference information in first column
\makeatletter
\def\@copyrightspace{\relax}
\makeatother

\AtBeginDocument{%
  \providecommand\BibTeX{{%
    \normalfont B\kern-0.5em{\scshape i\kern-0.25em b}\kern-0.8em\TeX}}}

\copyrightyear{2020}
\acmYear{2020}
\acmConference[Arxiv]{}
\acmBooktitle{}
\acmPrice{}
\acmDOI{}
\acmISBN{}

\usepackage{lipsum}
\usepackage{multirow}
\usepackage{graphicx}
\makeatletter
\newcommand{\printfnsymbol}[1]{%
  \textsuperscript{\@fnsymbol{#1}}%
}
\makeatother

\usepackage{breakurl}
\begin{document}
\sloppy
\title{Towards Detection of Subjective Bias using Contextualized Word Embeddings}

% \author{Anonymous Submission}

\author{Tanvi Dadu}
\affiliation{%
  \institution{NSIT Delhi}}
  \email{tanvid.co.16@nsit.net.in}
\authornotemark[1]

\author{Kartikey Pant}
\affiliation{%
  \institution{IIIT Hyderabad}}
\email{kartikey.pant@research.iiit.ac.in}
\authornote{The first two authors contributed equally to the work.}

\author{Radhika Mamidi}
\affiliation{%
  \institution{IIIT Hyderabad}}
\email{radhika.mamidi@iiit.ac.in}

% \affiliation{%

% \affiliation{%

% \affiliation{%



\renewcommand{\shortauthors}{Dadu, Pant, and Mamidi}
\begin{abstract}
Subjective bias detection is critical for applications like propaganda detection, content recommendation, sentiment analysis, and bias neutralization. This bias is introduced in natural language via inflammatory words and phrases, casting doubt over facts, and presupposing the truth. In this work, we perform comprehensive experiments for detecting subjective bias using BERT-based models on the Wiki Neutrality Corpus(WNC). The dataset consists of $360k$ labeled instances, from Wikipedia edits that remove various instances of the bias. We further propose BERT-based ensembles that outperform state-of-the-art methods like $BERT_{large}$ by a margin of $5.6$ F1 score. 
\end{abstract}

%To appear in Companion Proceedings of the Web Conference 2020 (WWW '20 Companion)

% \keywords{datasets, neural networks, gaze detection, text tagging}


\maketitle

\section{Introduction}
In natural language, subjectivity refers to the aspects of communication used to express opinions, evaluations, and speculations\cite{Wiebe1994}, often influenced by one's emotional state and viewpoints. Writers and editors of texts like news and textbooks try to avoid the use of biased language, yet subjective bias is pervasive in these texts. More than $56\%$ of Americans believe that news sources do not report the news objectively \footnote{\url{https://news.gallup.com/opinion/gallup/235796/americans-misinformation-bias-inaccuracy-news.aspx}}, thus implying the prevalence of the bias. Therefore, when presenting factual information, it becomes necessary to differentiate subjective language from objective language.

There has been considerable work on capturing subjectivity using text-classification models ranging from linguistic-feature-based models\cite{Recasens2013} to finetuned pre-trained word embeddings like BERT\cite{Jurafsky2020}. The detection of bias-inducing words in a Wikipedia statement was explored in \cite{Recasens2013}. The authors propose the "Neutral Point of View" (\textit{NPOV}) corpus made using Wikipedia revision history, containing Wikipedia edits that are specifically designed to remove subjective bias. They use logistic regression with linguistic features, including factive verbs, hedges, and subjective intensifiers to detect bias-inducing words. In \cite{Jurafsky2020}, the authors extend this work by mitigating subjective bias after detecting bias-inducing words using a BERT-based model. However, they primarily focused on detecting and mitigating subjective bias for single-word edits. We extend their work by incorporating multi-word edits by detecting bias at the sentence level. We further use their version of the \textit{NPOV} corpus called Wiki Neutrality Corpus(\textit{WNC}) for this work.

The task of detecting sentences containing subjective bias rather than individual words inducing the bias has been explored in \cite{hube2019neural}. However, they conduct majority of their experiments in controlled settings, limiting the type of articles from which the revisions were extracted. Their attempt to test their models in a general setting is dwarfed by the fact that they used revisions from a single Wikipedia article resulting in just $100$ instances to evaluate their proposed models robustly. Consequently, we perform our experiments in the complete \textit{WNC} corpus, which consists of $423,823$ revisions in Wikipedia marked by its editors over a period of 15 years, to simulate a more general setting for the bias.

In this work, we investigate the application of BERT-based models for the task of subjective language detection\footnote{Made available at \url{https://github.com/tanvidadu/Subjective-Bias-Detection}}. We explore various BERT-based models, including \textit{BERT}, \textit{RoBERTa}, \textit{ALBERT}, with their \textit{base} and \textit{large} specifications along with their native classifiers. We propose an ensemble model exploiting predictions from these models using multiple ensembling techniques. We show that our model outperforms the baselines by a margin of $5.6$ of F1 score and $5.95\%$ of Accuracy.

\section{Baselines and Approach}
In this section, we outline baseline models like $BERT_{large}$. We further propose three approaches: optimized BERT-based models, distilled pretrained models, and the use of ensemble methods for the task of subjectivity detection.

\subsection{Baselines}
% As baselines, we experiment with the following three models:
\begin{enumerate}
    \item \textit{\textbf{FastText}\cite{joulin2016tricks}}: It uses bag of words and bag of n-grams as features for text classification, capturing partial information about the local word order efficiently.
    \item \textit{\textbf{BiLSTM}}: Unlike feedforward neural networks, recurrent neural networks like BiLSTMs use memory based on history information to learn long-distance features and then predict the output. We use a two-layer BiLSTM architecture with GloVe word embeddings as a strong RNN baseline.
    % \item \textit{ULMFiT \cite{ulmfit2018}}: It is a language model pretrained on a large general-domain corpus. We fine-tuned the model using Discriminative finetuning and Slanted triangular learning rates on the training set.
    \item \textit{\textbf{BERT} \cite{Devlin2019}}: It is a contextualized word representation model that uses bidirectional transformers, pretrained on a large $3.3B$ word corpus. We use the $BERT_{large}$ model finetuned on the training dataset.
\end{enumerate}

\subsection{Proposed Approaches}
% We propose the following three approaches: 
\begin{enumerate}
    \item \textit{\textbf{Optimized BERT-based models}}: We use BERT-based models optimized as in \cite{2020roberta} and \cite{2020albert}, pretrained on a dataset as large as twelve times as compared to $BERT_{large}$, with bigger batches, and longer sequences. \textit{ALBERT}, introduced in \cite{2020albert}, uses factorized embedding parameterization and cross-layer parameter sharing for parameter reduction. These optimizations have led both the models to outperform $BERT_{large}$ in various benchmarking tests, like \textit{GLUE} for text classification and \textit{SQuAD} for Question Answering.
    \item \textit{\textbf{Distilled BERT-based models}}: Secondly, we propose to use distilled BERT-based models, as introduced in \cite{DistilBert}. They are smaller general-purpose language representation model, pre-trained by leveraging distillation knowledge. This results in significantly smaller and faster models with performance comparable to their undistilled versions. We finetune these pretrained distilled models on the training corpus to efficiently detect subjectivity.
    % They also use a combination of three losses: language modeling, distillation, and cosine-distance, to leverage the inductive biases learned during pre-training.
    \item \textit{\textbf{BERT-based ensemble models}}: Lastly, we use the weighted-average ensembling technique to exploit the predictions made by different variations of the above models. Ensembling methodology entails engendering a predictive model by utilizing predictions from multiple models in order to improve Accuracy and F1, decrease variance, and bias. We experiment with variations of $RoBERTa_{large}$, $ALBERT_{xxlarge.v2}$, $DistilRoBERTa$ and $BERT$ and outline selected combinations in \autoref{tab:experimental-results}.
\end{enumerate}

\section{Experiments}
\subsection{Dataset and Experimental Settings}
We perform our experiments on the \textit{WNC} dataset open-sourced by the authors of \cite{Jurafsky2020}.  It consists of aligned pre and post neutralized sentences made by Wikipedia editors under the neutral point of view. It contains $180k$ biased sentences, and their neutral counterparts crawled from $423,823$ Wikipedia revisions between $2004$ and $2019$. We randomly shuffled these sentences and split this dataset into two parts in a $90:10$ Train-Test split and perform the evaluation on the held-out test dataset. 

For all BERT-based models, we use a learning rate of $2*10^{-5}$, a maximum sequence length of $50$, and a weight decay of $0.01$ while finetuning the model. We use FastText's recently open-sourced automatic hyperparameter optimization functionality while training the model. For the BiLSTM baseline, we use a dropout of $0.05$ along with a recurrent dropout of $0.2$ in two $64$ unit sized stacked BiLSTMs, using softmax activation layer as the final dense layer.

\subsection{Experimental Results}
\autoref{tab:experimental-results} shows the performance of different models on the WNC corpus evaluated on the following four metrics: Precision, Recall, F1, and Accuracy. Our proposed methodology, the use of finetuned optimized BERT based models, and BERT-based ensemble models outperform the baselines for all the metrics.

Among the optimized BERT based models, $RoBERTa_{large}$ outperforms all other non-ensemble models and the baselines for all metrics. It further achieves a maximum recall of $0.681$ for all the proposed models. We note that DistillRoBERTa, a distilled model, performs competitively, achieving $69.69\%$ accuracy, and $0.672$ F1 score. This observation shows that distilled pretrained models can replace their undistilled counterparts in a low-computing environment.

We further observe that ensemble models perform better than optimized BERT-based models and distilled pretrained models. Our proposed ensemble comprising of $RoBERTa_{large}$, $ALBERT_{xxlarge.v2}$, $DistilRoBERTa$ and $BERT$ outperforms all the proposed models obtaining $0.704$ F1 score, $0.733$ precision, and $71.61\%$ Accuracy. 

\begin{table}[t]
\centering
\resizebox{0.475\textwidth}{!}{%
\begin{tabular}{|l|l|r|r|r|r|}
\hline
\textbf{} & \textbf{Models/Metrics} & \multicolumn{1}{l|}{\textbf{Precision}} & \multicolumn{1}{l|}{\textbf{Recall}} & \multicolumn{1}{l|}{\textbf{F1}} & \multicolumn{1}{l|}{\textbf{Acc}} \\ \hline
\multirow{3}{*}{\textbf{Baselines}} & FastText & 0.613 & 0.612 & 0.613 & 61.24\% \\ \cline{2-6} 
 & BiLSTM+GloVe  & 0.648 & 0.647 & 0.648 & 64.76\%  \\ \cline{2-6} 
 & $BERT_{large}$ & 0.681 & 0.587 & 0.631 & 65.66\% \\ \hline
\multirow{4}{*}{\textbf{Single Model}} & $ALBERT_{xxlarge.v2}$ & 0.667 & 0.579 & 0.620 & 64.56\% \\ \cline{2-6} 
 & DistillBERT & 0.731 & 0.608 & 0.664 & 69.28\% \\ \cline{2-6} 
 & DistillRoBERTa & 0.730 & 0.623 & 0.672 & 69.69\% \\ \cline{2-6} 
 & $RoBERTa_{large}$ & 0.723 & \textbf{0.681} & 0.702 & 71.09\% \\ \hline
\multirow{4}{*}{\textbf{Ensemble model}} & $BERT_{Ensemble}$+DistillBERT & 0.731 & 0.610 & 0.665 & 69.36\% \\ \cline{2-6} 
 & $RoBERTa_{Ensemble}$ & 0.732 & 0.679 & 0.704 & 71.57\% \\ \cline{2-6} 
 & RoBERTa+ALBERT & \multirow{2}{*}{\textbf{0.733}} & \multirow{2}{*}{0.677} & \multirow{2}{*}{\textbf{0.704}} & \multirow{2}{*}{\textbf{71.61\%}} \\
 & +DistillRoBERTa+BERT &  &  &  &  \\ \hline
\end{tabular}%
}
\caption{Experimental Results for the Subjectivity Detection Task}
\label{tab:experimental-results}
\end{table}

\section{Conclusion}
In this paper, we investigated BERT-based architectures for sentence level subjective bias detection. We perform our experiments on a general Wikipedia corpus consisting of more than $360k$ pre and post subjective bias neutralized sentences. We found our proposed architectures to outperform the existing baselines significantly. BERT-based ensemble consisting of \textit{RoBERTa}, \textit{ALBERT}, \textit{DistillRoBERTa}, and \textit{BERT} led to the highest F1 and Accuracy. In the future, we would like to explore document-level detection of subjective bias, multi-word mitigation of the bias, applications of detecting the bias in recommendation systems.

\bibliographystyle{ACM-Reference-Format}
\bibliography{bias-detection}


\begin{thebibliography}{9}

%%% the Web address.

\ifx \showCODEN    \undefined \def \showCODEN     #1{\unskip}     \fi
\ifx \showDOI      \undefined \def \showDOI       #1{#1}\fi
\ifx \showISBNx    \undefined \def \showISBNx     #1{\unskip}     \fi
\ifx \showISBNxiii \undefined \def \showISBNxiii  #1{\unskip}     \fi
\ifx \showISSN     \undefined \def \showISSN      #1{\unskip}     \fi
\ifx \showLCCN     \undefined \def \showLCCN      #1{\unskip}     \fi
\ifx \shownote     \undefined \def \shownote      #1{#1}          \fi
\ifx \showarticletitle \undefined \def \showarticletitle #1{#1}   \fi
\ifx \showURL      \undefined \def \showURL       {\relax}        \fi
% invisible to TeX
\providecommand\bibfield[2]{#2}
\providecommand\bibinfo[2]{#2}
\providecommand\natexlab[1]{#1}
\providecommand\showeprint[2][]{arXiv:#2}

\bibitem[\protect\citeauthoryear{Devlin, Chang, Lee, and Toutanova}{Devlin
  et~al\mbox{.}}{2018}]%
        {Devlin2019}
\bibfield{author}{\bibinfo{person}{Jacob Devlin}, \bibinfo{person}{Ming-Wei
  Chang}, \bibinfo{person}{Kenton Lee}, {and} \bibinfo{person}{Kristina
  Toutanova}.} \bibinfo{year}{2018}\natexlab{}.
\newblock \bibinfo{title}{BERT: Pre-training of Deep Bidirectional Transformers
  for Language Understanding}.
\newblock
\newblock
\urldef\tempurl%
\url{http://arxiv.org/abs/1810.04805}
\showURL{%
\tempurl}


\bibitem[\protect\citeauthoryear{Hube and Fetahu}{Hube and Fetahu}{2019}]%
        {hube2019neural}
\bibfield{author}{\bibinfo{person}{Christoph Hube} {and}
  \bibinfo{person}{Besnik Fetahu}.} \bibinfo{year}{2019}\natexlab{}.
\newblock \showarticletitle{Neural Based Statement Classification for Biased
  Language}. In \bibinfo{booktitle}{\emph{12th ACM International Conference on
  Web Search and Data Mining (WSDM)}}.
\newblock


\bibitem[\protect\citeauthoryear{Joulin, Grave, Bojanowski, and Mikolov}{Joulin
  et~al\mbox{.}}{2016}]%
        {joulin2016tricks}
\bibfield{author}{\bibinfo{person}{Armand Joulin}, \bibinfo{person}{Edouard
  Grave}, \bibinfo{person}{Piotr Bojanowski}, {and} \bibinfo{person}{Tomas
  Mikolov}.} \bibinfo{year}{2016}\natexlab{}.
\newblock \bibinfo{title}{Bag of Tricks for Efficient Text Classification}.
\newblock
\newblock
\urldef\tempurl%
\url{http://arxiv.org/abs/1607.01759}
\showURL{%
\tempurl}
\newblock
\shownote{cite arxiv:1607.01759.}


\bibitem[\protect\citeauthoryear{Pryzant, Martinez, Dass, Kurohashi, Jurafsky,
  and Yang}{Pryzant et~al\mbox{.}}{2019}]%
        {Jurafsky2020}
\bibfield{author}{\bibinfo{person}{Reid Pryzant},
  \bibinfo{person}{Richard~Diehl Martinez}, \bibinfo{person}{Nathan Dass},
  \bibinfo{person}{Sadao Kurohashi}, \bibinfo{person}{Dan Jurafsky}, {and}
  \bibinfo{person}{Diyi Yang}.} \bibinfo{year}{2019}\natexlab{}.
\newblock \bibinfo{title}{Automatically Neutralizing Subjective Bias in Text}.
\newblock
\newblock
\showeprint{arXiv:1911.09709}


\bibitem[\protect\citeauthoryear{Recasens, Danescu-Niculescu-Mizil, and
  Jurafsky}{Recasens et~al\mbox{.}}{2013}]%
        {Recasens2013}
\bibfield{author}{\bibinfo{person}{Marta Recasens}, \bibinfo{person}{Cristian
  Danescu-Niculescu-Mizil}, {and} \bibinfo{person}{Dan Jurafsky}.}
  \bibinfo{year}{2013}\natexlab{}.
\newblock \showarticletitle{Linguistic Models for Analyzing and Detecting
  Biased Language.}. In \bibinfo{booktitle}{\emph{ACL (1)}}.
  \bibinfo{publisher}{The Association for Computer Linguistics},
  \bibinfo{pages}{1650--1659}.
\newblock
\showISBNx{978-1-937284-50-3}
\urldef\tempurl%
\url{http://dblp.uni-trier.de/db/conf/acl/acl2013-1.html#RecasensDJ13}
\showURL{%
\tempurl}


\bibitem[\protect\citeauthoryear{Sanh, Debut, Chaumond, and Wolf}{Sanh
  et~al\mbox{.}}{2019}]%
        {DistilBert}
\bibfield{author}{\bibinfo{person}{Victor Sanh}, \bibinfo{person}{Lysandre
  Debut}, \bibinfo{person}{Julien Chaumond}, {and} \bibinfo{person}{Thomas
  Wolf}.} \bibinfo{year}{2019}\natexlab{}.
\newblock \bibinfo{title}{DistilBERT, a distilled version of BERT: smaller,
  faster, cheaper and lighter}.
\newblock
\newblock
\showeprint{arXiv:1910.01108}


\bibitem[\protect\citeauthoryear{Wiebe}{Wiebe}{2002}]%
        {Wiebe1994}
\bibfield{author}{\bibinfo{person}{Janyce Wiebe}.}
  \bibinfo{year}{2002}\natexlab{}.
\newblock \showarticletitle{Tracking Point of View in Narrative}.
\newblock \bibinfo{journal}{\emph{Computational Linguistics}}
  \bibinfo{volume}{20} (\bibinfo{date}{07} \bibinfo{year}{2002}).
\newblock


\bibitem[\protect\citeauthoryear{Yinhan~Liu}{Yinhan~Liu}{2020}]%
        {2020roberta}
\bibfield{author}{\bibinfo{person}{Naman Goyal Jingfei Du Mandar Joshi Danqi
  Chen Omer Levy Mike Lewis Luke Zettlemoyer Veselin~Stoyanov Yinhan~Liu,
  Myle~Ott}.} \bibinfo{year}{2020}\natexlab{}.
\newblock \showarticletitle{RoBERTa: A Robustly Optimized BERT Pretraining
  Approach}. In \bibinfo{booktitle}{\emph{Submitted to International Conference
  on Learning Representations}}.
\newblock
\urldef\tempurl%
\url{https://openreview.net/forum?id=SyxS0T4tvS}
\showURL{%
\tempurl}
\newblock
\shownote{under review.}


\bibitem[\protect\citeauthoryear{Zhenzhong~Lan}{Zhenzhong~Lan}{2020}]%
        {2020albert}
\bibfield{author}{\bibinfo{person}{Sebastian Goodman Kevin Gimpel Piyush Sharma
  Radu~Soricut Zhenzhong~Lan, Mingda~Chen}.} \bibinfo{year}{2020}\natexlab{}.
\newblock \showarticletitle{ALBERT: A Lite BERT for Self-supervised Learning of
  Language Representations}. In \bibinfo{booktitle}{\emph{Submitted to
  International Conference on Learning Representations}}.
\newblock
\urldef\tempurl%
\url{https://openreview.net/forum?id=H1eA7AEtvS}
\showURL{%
\tempurl}
\newblock
\shownote{under review.}


\end{thebibliography}
\end{document}